# ROBUST MULTI-CAMERA VIEW FACE RECOGNITION

Dakshina Ranjan Kisku
Department of Computer Science and Engineering
Dr. B. C. Roy Engineering College,
Durgapur - 713206
India
Hunny Mehrotra
Department of Computer Science and Engineering
National Institute of Technology Rourkela
Rourkela – 769008
India
Phalguni Gupta
Department of Computer Science and Engineering
Indian Institute of Technology Kanpur
Kanpur – 208016
India
Jamuna Kanta Sing
Department of Computer Science and Engineering
Jadavpur University
Kolkata – 700032,
India
drkisku@ieee.org; hunny04@gmail.com; pg@cse.iitk.ac.in; , jksing@ieee.org

**ABSTRACT**
This paper presents multi-appearance fusion of Principal Component Analysis (PCA) and generalization of Linear Discriminant Analysis (LDA) for multi-camera view offline face recognition (verification) system. The generalization of LDA has been extended to establish correlations between the face classes in the transformed representation and this is called canonical covariate. The proposed system uses Gabor filter banks for characterization of facial features by spatial frequency, spatial locality and orientation to make compensate to the variations of face instances occurred due to illumination, pose and facial expression changes. Convolution of Gabor filter bank to face images produces Gabor face representations with high dimensional feature vectors. PCA and canonical covariate are then applied on the Gabor face representations to reduce the high dimensional feature spaces into low dimensional Gabor eigenfaces and Gabor canonical faces. Reduced eigenface vector and canonical face vector are fused together using weighted mean fusion rule. Finally, support vector machines (SVM) have trained with augmented fused set of features and perform the recognition task. The system has been evaluated with UMIST face database consisting of multiview faces. The experimental results demonstrate the efficiency and robustness of the proposed system for multi-view face images with high recognition rates. Complexity analysis of the proposed system is also presented at the end of the experimental results.

## 1. Introduction

Biometric authentication systems primarily use in security scenarios such as in sensitive area surveillance and access control. On the use of authentication systems largely in public and private places for access control and security, face recognition/verification has attracted the attention of vision researchers. Several approaches have been proposed for face recognition based on 2D and 3D images [18], [19], [20], [21]. Identity verification of authentic persons by their multi-view faces is a real valued problem in machine vision research. Although, many face recognition systems with frontal view faces have extensively been studied [1], [2], [3], [18], [19]. However, in rotated multi-view face recognition system some difficulties occur due to non-linear representation in feature spaces. To minimize this limitation, a global representation approach to non-linear feature spaces is necessary. In addition, variations in facial expression, lighting conditions, occlusions, environment and unwanted noises, affine distortions, clutter, etc, may also give some bad impact on the overall accuracy of face recognition system. To ensure robust recognition of multi-view faces with high recognition rate, some strategies have been developed [22], [23]. However, multi-view rotated face recognition still is a versatile and most challenging authentication system in terms of different viable constraints.

Face recognition approaches can be divided into two approaches. One is multiview-based face recognition approach [1-3], [11-13], [16] and another is single-view based approach [20]. In the multiview-based approach, the training is done using multiview face images and a query image is assumed to be matched to one of the existing model whereas the single-view based approach uses a canonical head pose for recognition. Normally, with the multiview-based approach, one might have "view specific models" [11] which makes the recognition process more complicated and even more time consuming.

In face recognition algorithms, appearance-based approach uses holistic texture features and makes a reduced set feature vector that can be applied either on the whole-face or on the divided block in a face image. Some of the well known sub-space based face recognition techniques are based on PCA [4], LDA [5], ICA [7], Kernel PCA [8], eigensignatures [13] etc. Principal component analysis (PCA) is a very traditional feature representation technique that uses to reduce the high dimensionality feature space to the smaller intrinsic dimensionality of feature space (independent variables). Bartlett et al. [7] have proposed ICA based face representation and found that it is better than PCA when cosines metric has been used as the similarity measure. Yang [8] has used Kernel PCA for feature extraction and recognition to show that the Kernel Eigen-faces method outperforms the conventional Eigen-faces method.

In a PCA based 'eigenface' [4] representation, when a face image projects into a feature space, the correlation among the components is found to be zero. On the other hand, LDA [5] based face recognition algorithm nonlinearly maps the input feature space onto a high-dimensional feature space, where the face pattern is linearly distributed. In a LDA face representation, when a face image projects into a feature space the variability among the face vectors in the same class is minimized and the variability among the face vectors in the different class is maximized. However, these kernel methods give no guarantee that this feature set is sufficient for better classification. Generally, original or transformed high-dimensional dataset projected onto the sub-space has higher variance but the classification capacity may be poor. Due to this poor classification capacity, we can extend the present LDA approach to more realistic and higher classification capacity oriented feature space which we call canonical covariate. When canonical covariate [14], [24], [27] (canonical covariate is a generalization of LDA) is used, the projected dataset onto the lower sub-space shows lower variance, but classification rate is high. Canonical variate [14] is used to combine the class of each data item as well as the features, which are estimating good set of features. In a canonical covariate representation, when a face image is projected in the feature space where the variability among the face sub-spaces in the same class is minimized while this variability contains the class of each data item as well as good set of features, which are estimated to be minimized. On the other hand, the variability among the face sub-spaces in the different classes is maximized while the variability contains the class of each data item as well as the good set of features, which have just been estimated to maximize. In simple realization, principal components are the linear combinations of the original features that capture the most variance in the dataset and in contrast, canonical covariates are linear combinations of the original features that capture the most correlation between two sets of variables [24], [27].

Face images have many nonlinear characteristics that are not addressed by linear analysis methods such as variations in illumination, pose and expression. Due to these limitations and multiple factors, this paper proposes a fusion approach, which integrates principal components and canonical covariate [14], [24], [27] of Gabor [10-11] responses that construct strong Gabor-eigenface representation and Gabor-canonical covariate representation. These representations have constructed from multiview faces and have been combined together into a robust representation which can remove the drawback of accurate localization of facial landmarks.

Convolutions of 2D Gabor wavelet transform [10-11] and multiview faces have been performed and this convolution has produced a set of high dimensional Gabor responses. These high dimensional feature spaces have characterized by spatial frequency, spatial locality and orientation. Gabor face representations are encoded by PCA and canonical covariate techniques and reduce the high dimensional input feature spaces into holistic low dimensional sub-spaces. When the Gabor face responses have projected into the lower dimensional principal component sub-spaces, the sub-space representations are called Gabor eigenfaces and when the Gabor face responses have projected into the lower dimensional canonical covariate sub-spaces, they are called canonical covariate sub-spaces. Then, we have fused these two representations using the proposed weighted mean fusion scheme into a feature vector and the feature vector consists of distinctive and significant set of features that characterizes the variations among all the faces in the class. Also this fusion scheme exhibits the characteristics of the features which are found in the optimal directions of representations. For face classification and recognition task, SVM classifier has trained with fused feature vectors. The fusion of PCA and canonical covariate has significantly improved the recognition performance over the existing methods [15-17]. This has been possible due to maximal mutual information estimated in the lower dimensional sub-space while features are extracted using PCA and using canonical covariate. We have also observed that dimensionality reduction of original feature vectors using principal components and canonical covariate have been used together without loss of significant characteristics of original representations.

The paper is organized as follows. Next section introduces Gabor wavelet transform for representation of face images. In Section 3 we describe PCA and canonical covariate for dimensionality reduction of the higher dimensional Gabor faces which have employed for multiview faces. Section 4 proposes a weighted fusion scheme. Section 5 describes the SVM classifier in the context of the proposed system. In section 6, experimental results are discussed. Complexity analysis in terms of time is presented in Section 7 and concluding remarks are given in Section 8.

## 2. Face Characterization by Gabor Filters

Gabor wavelet has been extensively studied in biometrics, like face, fingerprint and palmprint. Due to its well representation capability, Gabor wavelet filter is a feature extraction tool for some pattern recognition and biometric applications. Fundamentally, 2D Gabor filter [10-11] defined as a linear filter whose impulse response function has defined as the multiplication of harmonic function and Gaussian function in which Gaussian function has modulated by a sinusoid function. In this regard, the convolution theorem states that the Fourier transform of a Gabor filter's impulse response is the convolution of the Fourier transform of the harmonic function and the Fourier transform of the Gaussian function. Gabor function is a non-orthogonal wavelet and it can be specified by the frequency of the sinusoid $\omega = 2\pi f$ and the standard deviations $\sigma_x$ and $\sigma_y$. The 2D Gabor wavelet filter can be defined as follows

$$g(x, y : f, \theta) = \exp(-\frac{1}{2}(\frac{P^2}{\sigma_x^2} + \frac{Q^2}{\sigma_y^2}))\cos(2\pi f P)) \quad (1)$$

$$P = (x\sin\theta + y\cos\theta)$$
$$Q = (x\cos\theta - y\sin\theta)$$

where $f$ is the frequency of the sinusoidal plane wave along the direction $\theta$ from the $x$-axis, $\sigma_x$ and $\sigma_y$ specify the Gaussian envelop along $x$-axis and along $y$-axis, respectively. This can be used to determine the bandwidth of the Gabor filter. For the sake of experiment, 500 dpi gray scale face image with the size of 200 × 200 has been used. Along with this, 40 spatial frequencies are used, with $f=\pi/2^i$, ($i=1,2,...,5$) and $\theta=k\pi/8$ ($k=1,2,...,8$). For Gabor face representation, face image has convolved with the Gabor filter bank for capturing substantial amount of variations among face images in spatial locations. Gabor filter bank with five frequencies and eight orientations have used for generation of 40 spatial frequencies and for Gabor face extraction. In practice, Gabor face representation is very long and the dimension of Gabor feature vector is prohibitively large.

The proposed technique has been used multiview face images for robust and invariant face recognition, in which any profile or frontal view of query face can be matched to the database face image for face verification. First the face images have convolved with the Gabor wavelet filters and the convolution has generated 40 spatial frequencies in the neighborhood regions of the current spatial pixel point. For the face image of size 200 × 200, 1760000 spatial frequencies have been generated. Infact, the huge dimension of Gabor responses could cause the performance to be degraded and matching would be slow.

## 3. Dimensionality Reduction of Gabor Spaces by PCA and Canonical Covariate

The aim of dimensionality reduction of high dimensional features is to obtain a reduced set of features that reflects the relevance of the original feature set. The Gabor wavelet feature representation originated with very high dimensional space. It is necessary to reduce the high dimensional feature space to a low dimensional representation by selecting relevant and important features from the feature space. In this proposed work, we have used PCA [4] and canonical covariate holistic appearance based techniques to select the significant features from the Gabor face responses and hence to reduce the high dimensional data. In order to extract discriminatory feature information from face images, both PCA and canonical covariate are applied to the face images. PCA is used to reduce the original dimension of features to a compressed one while canonical covariate is used to extract discriminatory feature information from face. The main focus of applying canonical covariate with PCA is to identify distinct features which would be useful for intra-class and inter-class distinction.

### 3.1 Eigenface Treatment to Gabor Responses

Gabor face representation reflects the feature space of high dimensional features. This higher dimensional feature set contains complementary information as well as redundant noisy data. Using appearance based approaches, face images can be reduced to a set of holistic faces, such as eigenfaces. By applying PCA, Gabor faces can be compressed into a set of Gabor eigenfaces which represents optimal directions for the best representation of the Gabor face responses in the mean squared error sense. Let the training set of Gabor faces be $G_1, G_2, G_3, ..., G_m$ and also let each $G_i$ be $d$-dimensional feature vector. Here, $d$ is equal to the number of feature points for each Gabor face while the face image is represented by Gabor responses with 5 different frequencies and 8 orientations.

The average Gabor face can be defined as

$$\Psi = \frac{1}{m}\sum_{i}^{m} G_i \qquad (2)$$

Now, each Gabor face differs from the average Gabor face by the following vector

$$\eta_i = G_i - \Psi; \qquad (3)$$

The large set of Gabor vectors is then subject to PCA to identify the set of $m$ orthogonal column vectors $c_i$ ($i=1,2,3,…,m$) and their related eigenvalues.

Principal component analysis is defined by the following transformation matrix equation

$$y_i = W^t G_i \qquad (4)$$

where $W$ denotes a transformation matrix from which the orthogonal column vectors and their associated eigenvalues have been computed. Then the covariance matrix would be defined as

$$CM = \frac{1}{m}\sum_{i=1}^{m} \eta_i \eta_i^t \qquad (5)$$

where $\eta_i^t$ denotes transpose of the transformation vector which represents difference between Gabor face from average Gabor face in Equation (5). In addition, $c_i$ can be computed as

$$c_i = \sum_{p=1}^{m} \Phi_p \eta_p \qquad (6)$$

where $\Phi_p$ is the $p^{th}$ largest eigen value of the matrix $CM$ for $i = 1,2,3,…,m$. The eigenfaces are chosen as $m'$ (where $m' < m$) vectors $c_k$ ($k = 1,2,3,…,m'$) that correspond to the largest $m'$ eigenvalues of the matrix $CM$. In the next phase, a new gabor face can be transformed into its gabor eigenface features by the $w_k = c_k^t (G-\psi)$ for $k = 1,2,3,…, m'$ and the weights form a feature set $FS = [w_1,w_2,w_3,…,w_{m'}]$ that describes the contribution of each eigenface in representing the input gabor face image.

### 3.2 Canonical Covariate Treatment to Gabor Spaces

In practice, canonical variate [14], [24], [27] has used to project a dataset onto the sub-space and it shows lower variance, but classification probability is very high. Canonical variates combine the class of each data item as well as the features which are estimating good set of features.

To construct canonical variate representations for gabor face responses, we assume a set of gabor face responses of $C$ classes. Each class contains $n_k$ Gabor responses and a Gabor response from the $k^{th}$ class is $g_{k,i}$ for $i \in \{1,2,…,n_k\}$. Also assume that the $C_j$ class has mean $\mu_j$ and there are $d$-dimensional features (each $g_i$ is of $d$-dimensional vectors). We can write $\bar{\mu}$ for the mean of the class means, that is

$$\bar{\mu} = \frac{1}{C}\sum_{j=1}^{C} \mu_j \qquad (7)$$

and

$$\beta = \frac{1}{C-1}\sum_{j=1}^{C}(\mu_j - \bar{\mu})(\mu_j - \bar{\mu})^T. \qquad (8)$$

where, $\beta$ denotes the variance of the class means. In the generalized case, we can assume that each class has the identical covariance $\Sigma$, and that has the full rank. In order to obtain a set of axes where the feature points are grouped into some clusters belonging to a particular class and the classes are distinct. This involves finding a set of features that maximizes the ratio of the separation (i.e., variance) between the class means to the variance within each class. The separation between the class means is typically referred to as the between-class ($C_b$) variance and the variance within a class is typically referred to as the within class variance ($C_w$).

Let us consider, each class has the identical covariance $\Sigma$, which is either known or estimated as

$$\Sigma = \frac{1}{N-1}\sum_{s=1}^{C}\left\{\sum_{i=1}^{n_c}(g_{s,i} - \mu_s)(g_{s,i} - \mu_s)^T\right\}. \qquad (9)$$

From the Equations (8) and (9), the unit eigenvectors of $UV$ can be defined as

$$UV = \sum{}^{-1}\beta = [ev_1, ev_2,…,ev_m] \qquad (10)$$

where each $ev_i$ ($ev_i \mid i = 1,2,3,…,m$) denotes the eigenvalue and the dimension $m$ denotes the number of the eigenvalues and $ev_1$ is the largest eigen-value that gives a set of linear Gabor features that best separates the class means. Projection onto the basis $\{ev_1,ev_2,…,ev_k\}$ provides the $k$-dimensional set of linear features that best separates the class means.

## 4. Fusion of Eigenface and Canonical Face Cues

In any face verification method, matching between database face and query face is performed by computing distance to query face from database face. Distance is then compared with either a global threshold or local thresholds and this take the decision of acceptance or rejection.

In appearance based methods it has been seen that, both PCA and LDA have a strong correlation of features. We can use this apparent correlation characteristic of PCA and LDA for face recognition. By integrating the representations of PCA and LDA into a single representation, the integrated representation captures the merits of both PCA and LDA representations. In a PCA representation, face image has projected into feature space where the correlation among the components is zero. On the other hand, LDA nonlinearly maps the input feature space onto a high-dimensional feature space where the face pattern is linearly distributed. The fusion of PCA and LDA would be a worth to face recognition and verification.

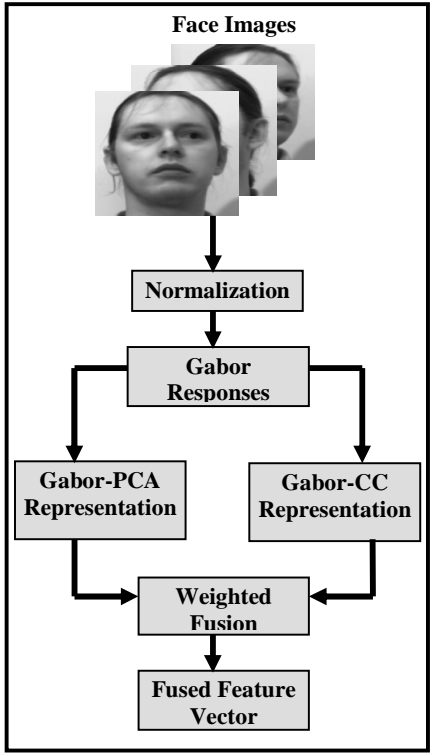

Figure 1.  Skeleton of the proposed fusion method.

In this paper, we have investigated the fusion scheme of PCA and LDA representations for 'multiview faces' in the context of face verification. Initially, Gabor face responses have projected in the PCA and canonical covariates sub-spaces which we have called Gabor-eigenfaces and Gabor-canonical covariate faces, respectively. The convolution of Gabor wavelet with face images has captured the minute characteristics in the neighbourhood regions of each pixel in spatially enhanced face. When PCA and canonical covariate are applied to high dimensional Gabor responses, it produces low dimensional feature vectors consisting of most relevant and spatially enhanced neighbourhood intensity information.

A fusion methodology for fusing two appearance-based approaches has been proposed by using PCA representation ("eigenface" representation) and canonical covariate representation. Fusion scheme of LDA and PCA in the context of the face recognition has already been applied [6], [15], [17]. Fig. 1 shows the skeleton of the proposed method. In this fusion, first the PCA and the canonical covariate representations of Gabor face responses are generated. The distance vectors for PCA and canonical covariate from all the faces in the database are then computed and we name these distance vectors as $\eta^{PCA}$ and $\eta^{CC}$, respectively. At the final step, these two vectors are combined to form a feature vector according to a proposed integration strategy.

The fusion scheme which has been proposed in this paper is characterized by 'weighted mean fusion rule'. According to the fusion scheme, a weight vector is computed from the separation of distributions of PCA and canonical covariate representations and these weights are then assigned to the integrated mean vectors.

Prior to fusion of these two representations, data normalization is performed of distance vectors in the interval of [0, 1] to reduce the range and large variability among the distance vectors. For normalization, simple 'min-max' normalization [36] rule is applied. It is assumed that, $Q$ is the dimension of each distance vector, where $\eta^{PCA} = [\eta_1^{PCA}, \eta_2^{PCA}, ..., \eta_Q^{PCA}]$ and $\eta^{CC} = [\eta_1^{CC}, \eta_2^{CC}, ..., \eta_Q^{CC}]$. If distance between a pair of data points is $d_i (d_i | i = 1, 2, ..., Q)$ for $\eta^{PCA}$ and $\eta^{CC}$, then the distance $d_i$ can be defined as the difference of the distributions originated from $\eta^{PCA}$ and $\eta^{CC}$. The separation of distributions can be defined as

$$d_i = \frac{\left\| \mu_i^{PCA} - \mu_i^{CC} \right\|}{\sqrt{(\sigma_i^{PCA})^2 + (\sigma_i^{CC})^2}} \quad (11)$$

where $\mu_i^{PCA}$ and $\mu_i^{CC}$ denote the means of the distance vectors $\eta^{PCA}$ and $\eta^{CC}$, respectively. Also, $\sigma_i^{PCA}$ and $\sigma_i^{CC}$ represent the standard deviations of $\eta^{PCA}$ and $\eta^{CC}$, respectively. Now, to compute the weights $w_i (w_i | i = 1, 2, ..., Q)$ which is proportional to $d_i$ can be defined as

$$w_i = \frac{d_i}{\sum_{i=1}^{Q} d_i} \quad (12)$$

for $1 \leq w_i \leq 1, \forall i, \sum_{i=1}^{Q} w_i = 1, \forall i$

In this weighted mean fusion rule, these weights are applied to individual separation computed from the corresponding data points of $\eta^{PCA}$ and $\eta^{CC}$. The weighted fusion strategy can be written as

$$F = \left[ w_1 \times \left( \frac{\eta_1^{PCA} + \eta_1^{CC}}{2} \right), w_2 \times \left( \frac{\eta_2^{PCA} + \eta_2^{CC}}{2} \right) \ldots w_Q \times \left( \frac{\eta_Q^{PCA} + \eta_Q^{CC}}{2} \right) \right] \quad (13)$$

In Equation (13), $F$ is the vector of fused mean values of dimension $Q$. Now these fused vectors are trained with support vector machines for classification.

## 5. SVM Classification

The proposed work uses support vector machines [9] to solve the problem of classifying faces. The training problem can be formulated as separating hyper-planes that maximizes the distance between closest points of the two classes. In practice, this is determined through solving quadratic problem. The SVM has a general form of the decision function for $N$ data points $\{x_i, y_i\}_{i=1}^{N}$, where $x_i \in \mathbf{R}^n$ the i-th input data is, and $y_i \in \{-1, +1\}$ is the label of the data. The SVM approach aims at finding a classifier of form:

$$y(x) = sign\left[ \sum_{i=1}^{N} \alpha_i y_i K(x_i, x) + b \right] \quad (14)$$

Where $\alpha_i$ are positive real constants and $b$ is a real constant, in general, $K(x_i, x) = \langle \phi(x_i), \phi(x) \rangle$ is known as kernel function where $\langle \bullet, \bullet \rangle$ inner product is, and $\phi(x)$ is the nonlinear map from original space to the high dimensional space. The kernel function can be various types. The linear function is defined as $K(x, y) = x \cdot y$; the Radial Basis Function (RBF) kernel function is $K(x, y) = \exp(-\frac{1}{2\sigma^2} \| x - y \|^2)$ and its polynomial kernel function is $K(x, y) = (x \cdot y + 1)^n$. In this experiment, SVM is used with two kernel functions, viz. linear function and Radial Basis Function (RBF). When the number of features is more, then it is not required to map the data to a higher dimensional feature space and in case of less number of features, then there is no need to map the features to a higher dimensional feature space.

Therefore, the linear kernel function with SVM is useful when the dimension of feature set is found to be large. On the other hand, RBF kernel nonlinearly maps samples into a higher dimensional space and can handle the case when the relation between class labels and attributes is nonlinear. Thus, RBF kernel can perform with less number of features.

Further, linear kernel can be considered as a special case of RBF kernel when the linear kernel with a penalty parameter has the same impact as RBF kernel for some parameters. The number of hyper-parameters has direct influence over the complexity of model selection. Linear kernel does not possess any hyper-parameters and due to this it is less complex than RBF kernel. However, on use of the linear kernel for large number of features increases the computational complexity. On the other hand, RBF kernel function having a hyper-parameter ($\sigma$) can be dealt with small number of features. Its complexity is found to be comparable with linear kernel and is found to be efficient one. In this experiment, since a reduced set of integrated features is used, RBF kernel is found to be more useful than linear kernel function.

SVM can be designed for either binary classification or multi-classification. For the sake of experiment, we have used binary classification approach. In binary classification, the goal of maximum margin is to separate the two classes by a hyperplane such that the distance to the support vectors is maximized. This hyperplane is known as the optimal separating hyperplane (OSH). For "one-vs-one" binary classification, the decision function (14) can be written as

$$f(x) = sign(\omega \cdot x - b) \quad (15)$$

where ω (inner product of weight vector) is obtained from the following equation

$$\omega = \sum_{i} y_i \alpha_i x_i \quad (16)$$

Here, the input feature vector x and weight vector ω determines the value of f(x). During classification, the input features with the largest weights correspond to the most discriminative and informative features. Therefore, the weights of the linear function can be used as final classification criterion for binary decision of multiview faces. For that, the pre-specified threshold is determined from Receiver Operating Characteristic (ROC) curves computed on an evaluation set by comparing it with the training set. The evaluation set is built from test dataset. Finally, this pre-specified threshold which is determined from evaluation set is used to compute different error rates on test set. ROC curves are produced by generating false acceptance and false rejection rates, and also the EER (Equal Error Rate) and recognition rate are computed separately.

## 6. Experimental Results and Discussion

In this section, the experimental results of the proposed system on UMIST face database [12-13], [26] for multiview face authentication have been analyzed. The face database consists of 564 face images of 20 distinct subjects. Faces in the database cover a mixed range of poses, races, sex and appearance such as different expressions, illuminations, wearing glasses or without glasses, having beard or without beard, different hair style etc. Some face images of a subject are shown in Fig. 2.

The experiment is accomplished in three steps. In the first step, Gabor wavelets are used and obtain feature representation of a face. Next step uses PCA and canonical covariate to reduce the high dimension of Gabor faces which contain significant linear features that separate classes efficiently. A weighted mean fusion rule is used to obtain the reduced faces from the PCA and canonical covariate representations of Gabor faces. Finally, classification of reduced faces is done by three classifiers, namely, K-Nearest Neighbour, Support Vector Machines (SVM) with two kernel functions, namely, linear function and Radial Basis Function (RBF).

The proposed system with three classifiers has been compared with PCA based and canonical covariate based multiview face recognition systems using SVM with RBF as classifier. Receiver Operating Characteristic (ROC) curves are shown in Fig. 3 while various error rates along with the recognition rates for all five systems are given in Table 1. It has been observed that the proposed fusion scheme which uses SVM classifier with RBF kernel function has achieved the accuracy of more than 98% and is the best among five systems.

The proposed system has been compared with the available well known systems in [15-17]. It has been observed that the system based on weighted fusion of PCA and canonical covariate is found to be more robust and reliable compared to [15-17].

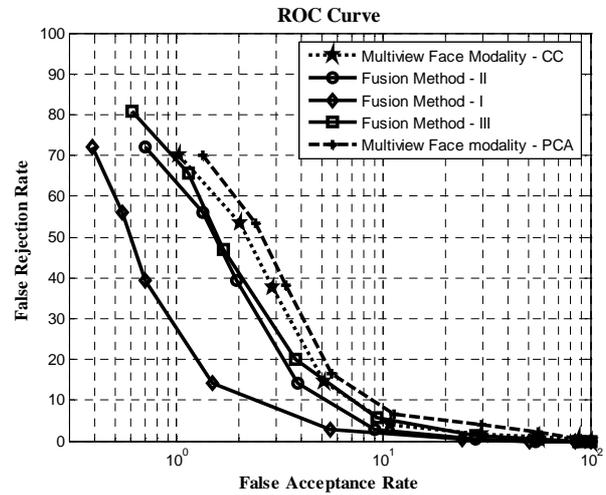

Figure 3. ROC curve for different methods

Table 1. Different Error Rates and Recognition Rates

| Error parameters → <br> Methods ↓ | FRR | FAR | Recognition Rate | EER |
|---|---|---|---|---|
| Multiview face recognition – PCA | 6.902% | 10.107% | 91.4955% | 8.5045% |
| Multiview face recognition – CC | 4.721% | 8.983% | 93.148% | 6.852% |
| Method - III (KNN – Classifier) | 4.402% | 7.4332% | 94.0828% | 5.9172% |
| Method - II (SVM-Linear) | 1.033% | 5.9014% | 96.5326% | 3.4671% |
| Method - I (SVM-RBF) | **1.098%** | **2.5168%** | **98.1926%** | **1.8074%** |

Further, experimental results reveal that the proposed system is computationally more efficient and the representation of faces based on Gabor wavelets are much more precise and can capture detail information. Further, the system making use of weighted fusion of PCA and canonical covariates for dimensionality reduction and of SVM for classification becomes a state-of-art recognition system for multiview faces.

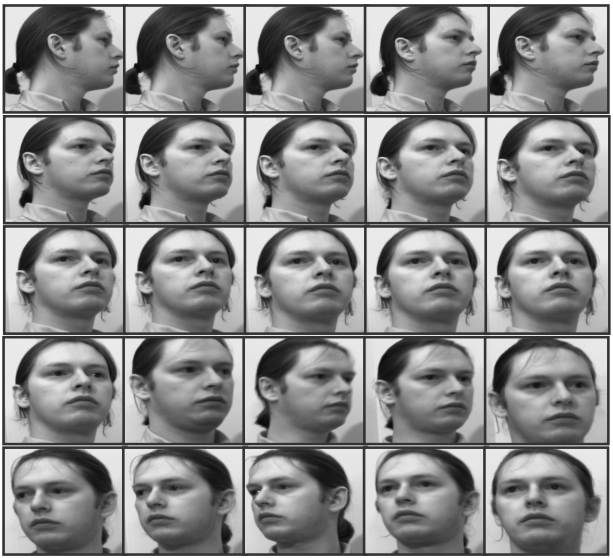

Figure 2. Face images of a subject from UMIST face database

## 7. Complexity Analysis

It is essential to compute time complexity of the proposed system to determine its efficiency. Computational complexity reveals the requirements of different computational resources which include hardware requirements and time required for execution of the algorithm. In the proposed multiview face recognition system, different phases require different computing resources. However, apart from hardware requirements, time complexity requirement is considered, in particular in this analysis. The proposed system is divided into several steps, such as Gabor wavelet coefficients extraction, Gabor eigenface and Gabor canonical face computation for obtaining prominent set of features. Later weighted mean fusion rule is applied to fuse the Gabor eigenface representation and Gabor canonical face representation into a weighted feature set. Finally, Support Vector Machines are used for classification of faces.

In order to compute the total time required by the proposed system, we start from face image acquisition and pre-processing operations. Let, time required to acquire a face image be $T1$ and for pre-processing $T2$ time is required. These two operations require constant time. Since for each face image acquisition constant time is required in a controlled environment and also for pre-processing operations include image enhancement and face localization constant time is required. Therefore, we can write,

Time for face image acquisition, $T(A) = T1\ (constant)$,
Time for pre-processing, $T(P) = T2\ (constant)$,

where, $T$ denotes time taken in step.

In order to extract wavelet coefficients from a face image, Gabor wavelet is convolved with a face image of dimension 200×200 and 40 spatial frequencies are generated for a pixel point. Therefore, the time required to generate 200×200×40 spatial frequencies would be $O(n^2)$. So, we can write,

Time taken by wavelet coefficients extraction = $O(n^2)*C1$.

In the next step, PCA is applied to Gabor responses of $C1*n^2$ dimensions. Therefore, from Equation (6), we can transform Gabor face into Gabor eigenface features by $w_k = c^t_k\ (G-\psi)$ for $k = 1,2,3,…, m'$ and the weights form a feature set $FS = [w_1,w_2,w_3,…,w_{m'}]$ which best describes the contribution of each eigenface in representing the input Gabor face image. $c^t_k$ takes $O(x)$ (here, $x<n$, x be the dimension of $c^t_k$) time and the remaining part $(G-\psi)$ will take $O(n^2)$ time. So, total time requires generating Gabor eigenface would be as follows

$T(w_k)= O(x)*O(n^2) \approx O(n^2)$;

Similarly, when canonical covariate is applied to Gabor face responses, time taken to generate identical covariance matrix (from Equation (9)) would be as follows

$$T(\Sigma) = \frac{1}{N-1}\sum_{s=1}^{C}\left\{\sum_{i=1}^{n_c}(g_{s,i}-\mu_s)(g_{s,i}-\mu_s)^T\right\}.$$
$= 1/O(n)*O(n^4)$
$\approx O(n^3)$

Since, number of classes, i.e., $\sum_{s=1}^{C}$ is constant and the term inside the inner summation takes $O(n^4)$ time.

From Equation (10), the eigenvector $UV$ takes the time which would be multiplied by the time taken by variance of class means, i.e., $\beta$ to $T(\Sigma)$ can be written for $UV$ be as follows

$T(UV) = O(n^3)*O(n)$
$\approx O(n^4)$,

where $\beta$ takes $O(n)$ time.

While fusion of these two representations is performed, viz. fusion of Gabor eigenface and Gabor canonical face, time to compute to generate weighted feature set would be

$T(F) = T(w_k) + T(UV)$,

$= O(n^2)+ O(n^4)*C$,
$\approx O(n^4)$

where $C$ be the weight which is constant in Equation (13).

Finally, three classifiers are used to classify the faces. Such as KNN, SVM with linear kernel and SVM with RBF kernel function.

When KNN is used as classifier, $O(n)$ time will be required and when SVM with linear kernel function is used, time taken by the classifier found from Equation (16) would be $O(n^2)$. Since, $y$ and $\omega$ are both constants, and $x$ is a feature vector which is directly related to linear kernel function. On the other hand, when SVM with RBF kernel is used, time taken to classify a face can be $O(n^3)$.

Now, the total time required by each classifier along with the other steps would be as follows,

(a) Total time required by the proposed system (when KNN classifier is used) $T(KNN) = T1+T2+T(F)+O(n)$,

Therefore,

$T(KNN) = T1+T2+O(n^4)+O(n) \approx O(n^4)$

(b) Total time required by the proposed system (when linear kernel is used with SVM) $T(Linear) = T1+T2+T(F)+O(n^2)$;

Therefore,
$T(Linear) = T1+T2+O(n^4)+O(n^2) \approx O(n^4)$;

(c) Total time required by the proposed system (when SVM is used with RBF kernel) $T(RBF) = T1+T2+T(F)+O(n^3)$,

Therefore,
$T(RBF) = T1+T2+O(n^4)+O(n^3) \approx O(n^4)$;

Time complexity analysis shows that the time taken by the three classifiers is approximately identical as $O(n^4)$. However, their performances vary with certain constraints discussed in the previous sections.

## 8. Conclusion

This paper has proposed a novel and robust face recognition method which can handle pose, illumination, occlusion, expression problems efficiently. The method is based on the Gabor wavelet representations of multiple views of faces. Due to high dimensionality of Gabor face responses, PCA and Canonical Covariate have been applied to obtain the reduced set of features. SVM has been used to classify the faces in binary classification pattern. A weighting fusion strategy has been proposed to fuse the reduced sets of features. PCA and canonical covariate are combined with the weighted mean fusion-based combination rule and the performance of RBF kernel based SVM is found to be much better than that of the linear kernel based SVM classifier and the K-nearest neighbour based classifier. The proposed system has been tested on UMIST database of multiview faces. The ROC curves show the robustness and reliability of the recognition system with the accuracy of more than 98%.

## Acknowledgement

Authors like to thank reviewers for their valuable comments which have helped to improve the quality of the paper.

## References

[1] J. Y. Gann, and Y. W. Zhang,, A new approach for face recognition based on singular value features and neural networks, *Acta Electronica Sinica, 32(1),* 2004, 56 – 58.

[2] J. Y. Gann, Y. W. Zhang, and S. Y. Mao, Adaptive principal components extraction algorithm and its applications in the feature extraction of human face, *Acta Electronica Silica, 30(7),* 2002, 1013 – 1016.

[3] Y. W. Zhang and J. Y. Gann, Human-Computer nature interaction, *Beijing National Defence Industry Press, 2004,* 20 – 85.

[4] M. Turk, and A. Pentland, Eigenfaces for recognition, *Journal of Cognitive Neuroscience, 3(1),* 1991, 71 – 86.

[5] P. Belhumeur, J. Hespanha, and D. Kriegman, Eigenfaces vs. fisherfaces: Recognition using class specific linear projection, *Proceedings of the Fourth European Conference on Computer Vision,* 1, 1996, 45 – 58.

[6] M. A. Grudin, On internal representations in face recognition systems, *Pattern Recognition, 33(7),* 2000, 1161 – 1177.

[7] M. S. Bartlett, J. R. Movellan, and T. J. Sejnowski, Face recognition by independent component analysis, *IEEE Transactions on Neural Networks, 13(6),* 2002, 1450 – 1464.

[8] M. H. Yang, Kernel Eigenfaces vs. kernel fisherfaces: Face recognition using kernel methods," *Proceedings of IEEE International Conference on Automatic Face and Gesture Recognition,* 2002, 215 – 220.

[9] C. J. C. Burges, A tutorial on support vector machines for pattern recognition, *Data Mining and Knowledge Discovery, 2(2),* 1998, 121–167.

[10] T. S. Lee, Image representation using 2D gabor wavelets, *IEEE Transactions on Pattern Analysis and Machine Intelligence, 18,* 1996, 959 – 971.

[11] M. J. Lyons, S. Akamatsu, M. Kamachi, and J. Gyoba, Coding facial expressions with gabor wavelets, *Proceedings of the IEEE International Conference on Automatic Face and Gesture Recognition,* 1998, 200 – 205.

[12] http://images.ee.umist.ac.uk/danny/database.html.

[13] D. Graham and N. Allinson, Characterizing virtual eigensignatures for general purpose face recognition, *Face Recognition: From Theory to Applications, ser. NATO ASI Series F, Computer and Systems Sciences, 163,* 1998, 446 – 456.

[14] H. Hotelling, Relations between two sets of variates, *Biometrika, 28,* 1936, 321–377.

[15] F. Roli and J. Kittler Eds., Proceedings of the Third International Workshop on Multiple Classifier Systems, Springer LNCS 2364, 2002.

[16] B. Achermann and H. Bunke, *Combination of Classifiers on the Decision Level for Face Recognition*, Technical Report IAM-96-002, Institut für Informatik und angewandte Mathematik, Universität Bern, 1996.

[17] G.L. Marcialis and F.Roli, Fusion of LDA and PCA for face recognition, *AI*IA Workshop on Machine Vision and Perception, held in the context of the 8th Meeting of the Italian Association for Artificial Intelligence*, 2002, 10 – 13.

[18] W. Zhao, R. Chellappa, A. Rosenfeld, J. Phillips, Face recognition: A literature survey, *ACM Computing Surveys, 12,* 2003, 399-458.

[19] F. S. Samaria, *Face recognition using hidden markov models*, PhD thesis, University of Cambridge, 1994.

[20] L. Wiskott, J. Fellous, N. Kruger, and C. Malsburg, Face recognition by elastic bunch graph matching, *IEEE Transactions on Pattern Analysis and Machine Intelligence, 19(7),* 1997, 775-779.

[21] C. Beumier, and M. P. Acheroy, Automatic face authentication from 3D surface, *Proceedings of the British Machine Vision Conference*, 1998, 449-458.

[22] Z. G. Fan, B. L. Lu, Multi-view face recognition with min-max modular SVMs, *Proceedings of the ICNC*, Lecture Notes in Computer Science, 3611, 2005, 396-399.

[23] Z. G. Fan, B. L. Lu, Fast recognition of multi-view faces with feature selection, *10th IEEE International Conference on Computer Vision (ICCV'05)*, 1, 2005, 76-81.

[24] B. S. Everitt, and G. Dunn, *Applied multivariate data analysis* (New York: Oxford University Press, 1992).

[25] A. A. Ross, *Handbook of multibiometrics* (Springer, 2006).

[26] A. Rattani, D. R. Kisku, A. Logario, and M. Tistarelli, Facial template synthesis based on SIFT features, Proceedings of the IEEE Workshop on Automatic Identification Advanced Technologies, 2007, 69 – 73.

[27] D. A. Forsyth, and J. Ponce, *Computer vision: A modern approach* (Pearson Education, Second Edition, 2005).